\documentclass{article}
\usepackage{spconf,amsmath,graphicx}
\usepackage{multirow}
\usepackage{booktabs}
\usepackage{bbding}
\usepackage{pifont}
\usepackage{wasysym}
\usepackage{amssymb}
\usepackage{multicol}
\usepackage{multirow}
\usepackage{makecell}
\usepackage{setspace}
\usepackage{bm}
\usepackage{caption}
\usepackage{balance}

\title{CLIP-based Synergistic Knowledge Transfer for Text-based Person Retrieval}
\name{Yating Liu\textsuperscript{12},
Yaowei Li\textsuperscript{32},
Zimo Liu\textsuperscript{2},
Wenming Yang\textsuperscript{1},
Yaowei Wang\textsuperscript{2*},
Qingmin Liao\textsuperscript{1*}\thanks{*Corresponding author 
}
}

\address{\textsuperscript{1} Shenzhen International Graduate School, Tsinghua University, China \\
\textsuperscript{2} Peng Cheng Laboratory,  China \\
\textsuperscript{3} School of ECE, Peking University, China \\
}

\begin{document}
%
\maketitle
\begin{abstract}
Text-based Person Retrieval (TPR) aims to retrieve the target person images given a textual query.
The primary challenge lies in bridging the substantial gap between vision and language modalities, especially when dealing with limited large-scale datasets. 
In this paper, we introduce a \emph{CLIP-based Synergistic Knowledge Transfer} (CSKT) approach for TPR.
Specifically, to explore the CLIP's knowledge on input side, we first propose a \emph{Bidirectional Prompts Transferring} (BPT) module constructed by text-to-image and image-to-text bidirectional prompts and coupling projections.
Secondly, Dual Adapters Transferring (DAT) is designed to transfer knowledge on output side of Multi-Head Attention (MHA) in vision and language. 
This synergistic two-way collaborative mechanism promotes the early-stage feature fusion and efficiently exploits the existing knowledge of CLIP.
CSKT outperforms the state-of-the-art approaches across three benchmark datasets when the training parameters merely account for 7.4\% of the entire model, demonstrating its remarkable efficiency, effectiveness and generalization.

\end{abstract}
\section{Introduction}
\label{sec:intro}
Text-based Person Retrieval (TPR) \cite{Li_2017_CVPR} refers to retrieving images of individuals from a large-scale image gallery based on textual descriptions. 
The core challenge is to establish cross-modal alignments between two distinct feature spaces (vision and language) in the absence of fine-grained image-text datasets of pedestrians.

Existing TPR methods are typically categorized into two types:
\emph{global-matching}  and \emph{local-matching} methods. 
\emph{Global-matching} \cite{han2021text,shu2022see,jiang2023cross,zhang2018deep,zheng2020dual} aligns images and texts into a shared latent embedding space by directly calculating the global features using specific loss functions. 
However, these methods often overlook crucial local features related to visual cues and textual attributes, which are vital for fine-grained tasks.
To address this limitation, \emph{local-matching} methods \cite{shao2022learning, yan2022clip, chen2022tipcb,wang2020vitaa, ding2021semantically} have been introduced to delve into local features.
They explicitly segment person images into stripes or patches, and analyze texts with additional parsers.
Moreover, implicit local alignments can be achieved through cross-modal attention or MASK mechanisms.
While \emph{local-matching} methods bring improved performance, their computational overhead during training and inference due to the extra complex modules is an obstacle when deployed in real-time applications.

Recently, Visual-Language  (V-L) large models \cite{li2022blip,clip} have drawn significant attention due to their remarkable capacity to acquire rich and robust semantic representations.  
Among these, Contrastive Language Image Pre-training (CLIP) \cite{clip}, which is trained on a web-scale open-vocabulary dataset comprising 400 million text-image pairs, is most widely utilized for downstream tasks.
However, full fine-tuning carries the risk of damaging the existing knowledge of CLIP and may lead to overfitting on specific downstream tasks.
To address this concern, Parameter-Efficient Transfer Learning (PETL) methods \cite{chen2022adaptformer, khattak2023maple, gao2021clip, zhou2022learning} have been proposed recently, which exhibit strong performance in exploiting the original knowledge of pre-trained models while only fine-tuning a few  parameters, such as prompt tuning and adapter tuning,  \emph{etc}.
Generally, most PETL methods belong to \emph{independent} tuning, where they fine-tune uni-modal or multi-modal prompts separately, often neglecting inter-modal prompt fusion. 
Furthermore, a \emph{dependent} approach called Multi-modal Prompt Learning (MaPLe) \cite{khattak2023maple} is introduced to enable cross-modal prompt fusion, which establishes the directional connections between different types of prompts.
\begin{figure*}[!ht]
  \centerline{\includegraphics[scale=0.23]{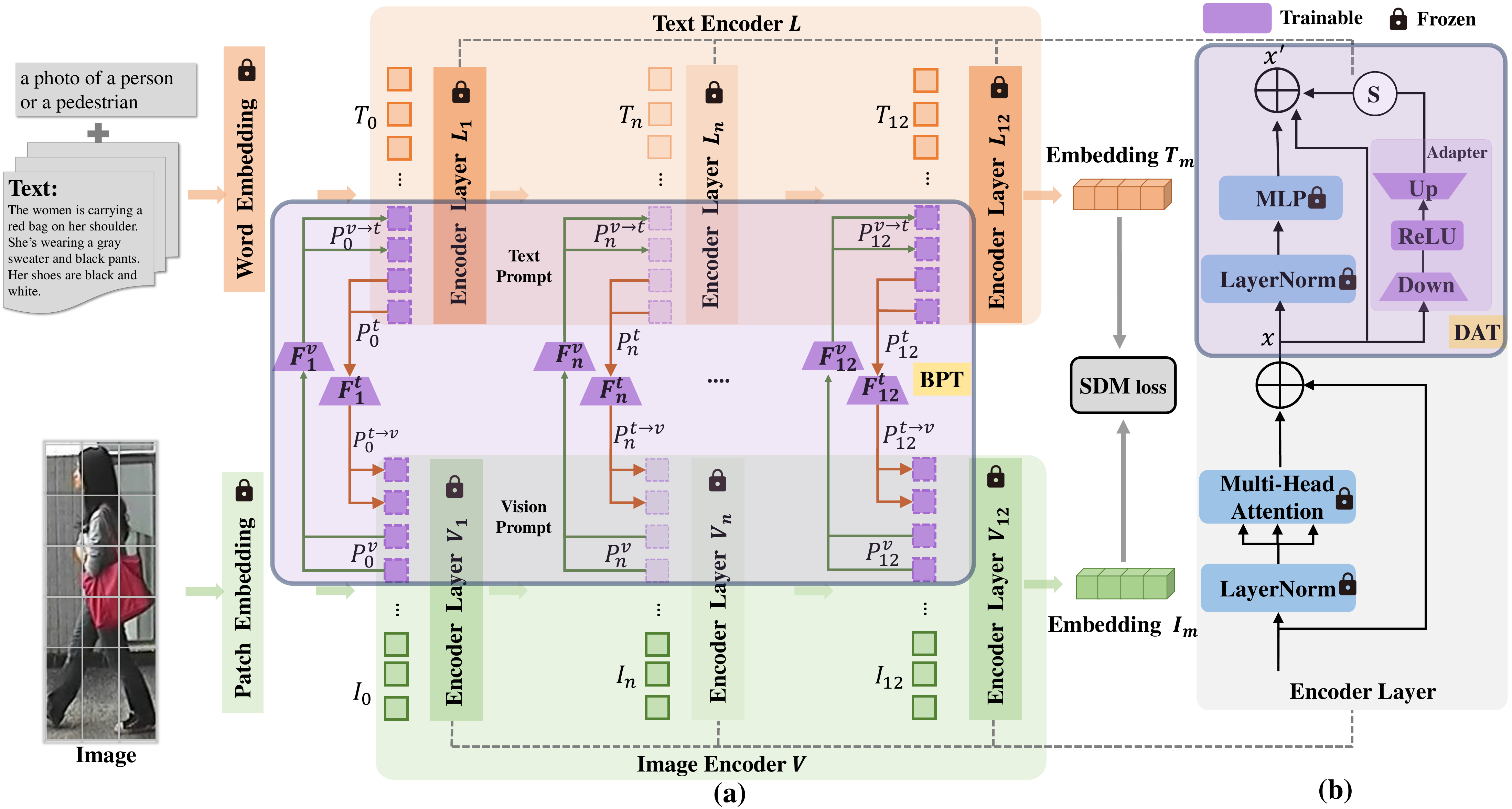}}
  \vspace{-2mm}
  \caption{\textbf{Overview of the proposed CSKT framework.}
 The main backbone of CSKT is depicted in Fig (a), which consists of dual encoders (image and text),  and two transfer learning modules (purple components), \emph{i.e.}, BPT and DAT. 
 The specific details of DAT are shown in Fig (b), which includes the adapted MLP of encoder layer in two branches. The two dashed lines represent the inside of the two branches. Only the parameters corresponding to the purple components are \emph{trainable}, while others are \emph{forzen}. The SDM depicted in Section 2.4 is utilized as a unique training loss.}
  \label{fig:framework}
\vspace{-3mm}
\end{figure*}

Motivated by the above observation, we propose a cross-Modal transferring method called \emph{CLIP-based Synergistic Knowledge Transfer} (CSKT), where two novel transfer modules including Bidirectional Prompts Transferring (BPT) and Dual Adapters Transferring (DAT) are designed to  do knowledge transferring on different locations  within CLIP.
Our contributions can be summarized as follows:

\textbf{(1)} We propose a novel parameter-efficient transfer learning (PETL) framework CSKT. To the best of our knowledge, it is the first time to achieve PETL for TPR task.

\textbf{(2)}  The synergistic two-way collaborative mechanism established by BPT and DAT, collaboratively enhances the deep fusion of V-L feature representations, and thoroughly leverages the CLIP's underlying capacities in rich knowledge and cross-modal alignment.

\textbf{(3)} Extensive experiments on CUHK-PEDES, ICFG-PEDES and RTSPReid datasets demonstrate the competitive performance of CSKT. Remarkably, only \textbf{12M} parameters (7.4\% of the total) are trainable in CSKT, resulting in significant reductions in computational and storage costs.  

\section{METHOD}
\label{sec:method}
Our method are elaborated by introducing modal architecture in Section~\ref{ssec:Framework} and two transferring modules in Section~\ref{ssec:BPT} and ~\ref{ssec:DAT}. CSKT attains strong knowledge transfer learning abilities by only fine-tuning a small number of parameters while keeping the backbone of CLIP frozen.

\subsection{Framework}
\label{ssec:Framework}
The overall architecture of our proposed CSKT framework is shown in Figure \ref{fig:framework}, which consists of two unimodal encoders (image and text) and two transferring modules,  \emph{i.e.}, Bidirectional  Prompts Transferring (BPT) and Dual Adapters Transferring (DAT).
Specifically, We adopt 12-layer transformer blocks of CLIP for both image and text encoders, respectively.
BPT makes the context prompts learnable in both vision and language branches by bidirectional prompt projections including text-to-image and image-to-text prompt projections.
Moreover, DAT inserts dual adapters into encoder layers of two branches to replace primitive MLP in MHA architecture.

For an image-text pair, we first concatenate initial prompts and raw text as input and feed it into text encoder to obtain a sequence of textual representations $\{t_{bos}, t_1,\cdots,t_N, t_{eos}\}$, where $t_{eos}$ is the global textual representation, $t_{bos}$ is the start representation, and the other is the token representation.
Simultaneously, we input the image patches (vision prompts and raw image patches) into image encoder to obtain visual representations $\{v_{cls}, v_1,\cdots,v_M\}$ with $v_{cls}$ being the global visual representation, and $v_i$ $(i=1,\cdots,M)$ being the patch representation.
The visual and textual representations are finally interacted and calculated by Similarity Distribution Matching (SDM) \cite{jiang2023cross}  which is an improved loss function across different modalities. 

\subsection{Bidirectional Prompts Transferring}
\label{ssec:BPT}
To  explore the early-stage feature fusion representations and to mine the potential of large model (CLIP) in TPR, inspired by MaPLe\cite{jiang2023cross}, we design bidirectional learnable prompt tokens and multiple learnable coupling layers within vision and language branches, enabling a deep interplay of cross-modal.


\noindent\textbf{Bidirectional Deep Language Prompting.} 
In the textual modality, bidirectional text prompts are composed of single-generated  and vision-to-language prompts.
Concretely, the input word embeddings are defined as $\left[ {P_0^t,P_0^{v \to t},{T_0}} \right]$, where $P_0^t \in {\mathbb{R}^{C \times D_t}}$ indicates the first $C$ embeddings of prompt text ``\emph{a photo of a person or a pedestrian}'', $P_0^{v \to t} \in {\mathbb{R}^{C \times D_t}}$ is generated by vision-to-language projection, and ${T_0} = [{T_0}^1,{T_0}^2,...,{T^m}]$ represents embeddings of other fixed tokens. The subsequent transformer layers concatenate the output from previous layer ${T_{i - 1}}$ and new learnable prompts $\left[{P_{i - 1}^t,P_{i - 1}^{v \to t}}\right]$ as the middle-layer input. The $J$ encoder layers are expressed as:
\vspace{-1mm}
\begin{equation}
\left[ {\_,\_,{T_i}} \right] = {L_i}\left( {\left[ {P_{i - 1}^t,P_{i - 1}^{v \to t},{T_{i - 1}}} \right]} \right)\begin{array}{*{5}{c}}
{}&{i = 1,2,...,J}
\end{array}
\end{equation}

\noindent\textbf{Bidirectional Deep Vision Prompting.}
Similarly, vision prompts are constructed by single-side vision prompts and language-to-vision prompts generated from  $J$ coupling layers in the other direction. 
Specifically, the input image embeddings are represented as $\left[ {P_0^v,P_0^{t \to v},{I_0}} \right]$, where initial vision prompts $P_0^t \in {\mathbb{R}^{C \times D_v}}$ are random vectors,
$P_0^{t \to v} \in {\mathbb{R}^{C \times D_v}}$ is the coupled representation, and ${I_0}$ indicates other frozen tokens. 
The $J$ image encoder layers are formulated as:
\vspace{-1mm}
\begin{equation}
\left[ {I_i},{\_,\_} \right] = {V_i}\left( {\left[ {I_{i - 1}},P_{i - 1}^v,P_{i - 1}^{t \to v} \right]} \right)\begin{array}{*{20}{c}}
{}&{i = 1,2,...,J}
\end{array}
\end{equation}

\noindent\textbf{Bidirectional Cross-modal Prompt Coupling.}
To promote a robust mutual synergy, coupling functions establish a bridge between two branches with bidirectional linear layers. $P_i^{t \to v} = F_i^t\left( {P_i^t} \right)$ is the $i$th function which maps the $D_t$ dimensional language inputs to $D_v$ vision outputs, and vice versa, $P_i^{v \to t} = F_i^v\left( {P_i^v} \right) $ is the projection in another direction.

\subsection{Dual Adapters Transferring}
\label{ssec:DAT}
To further straightforwardly mine original knowledge of two branches on the feature output side of MHA, dual adapters are incorporated into MLP of two encoders (Figure~\ref{fig:framework}(b)), which is the bottleneck structure including a down-projection layer, an up-projection layer and a ReLU layer between the two projection layers, formally via:
\vspace{-1mm}
\begin{equation}
{x_{out}} = x + {x_{MLP}} + s \cdot {\rm{Re}}LU\left( {LN\left( x \right) \cdot {W_d}} \right) \cdot {W_u}
\vspace{-2.5mm}
\end{equation}

\subsection{Train and Inference}
\label{ssec:train_inference}
In \emph{training} phase, we only make the newly added parameters (bidirectional V-L prompts and adapters) fine-tuned and keep the original weights of CLIP frozen. SDM loss \cite{jiang2023cross} is introduced to calculate the KL divergence between image-text similarity score distributions of V-L embeddings and true normalized labels distributions. 
We employ the bidirectional SDM loss including image-to-text and text-to-image, which is formulated as:
\vspace{-1mm}
\begin{equation}
\mathcal{L} = {\mathcal{L}_{sdm}} = {\mathcal{L}_{_{i2t}}} + {\mathcal{L}_{_{t2i}}}
\end{equation}

During \emph{inference}, similarities between text and image embeddings are calculated by the trained network, which contain bidirectional prompts. 
we further process the top-k candidates for each query and obtain relevant evaluation criteria.

\section{EXPERIMENTS}
\label{sec:pagestyle}

\subsection{Datasets and Implementation Details}
\label{ssec: datasets}

Extensive experiments are conducted on three official datasets.
\textbf{CUHK-PEDES} contains 40,206 images and 80,412 textual descriptions for 13,003 identities. The training set consists of 11,003 identities, and validation set and test set have 1,000 identities.
\textbf{ICFG-PEDES} contains 54,522 images for 4,102 identities. Each image corresponds to one description. The training and test sets contain 3,102 and 1,000 identities respectively.
\textbf{RSTPReid} contains 20505 images of 4,101 identities. Each image has 2 descriptions. The training, validation and test sets contain 3701, 200 and 200 identities respectively.

The pre-trained model CLIP is adopted as backbone, consisting of 12 V-L encoder layers. 
The image is resized to 384 × 128, and the textual token sequence is fixed at 77 tokens.
CSKT is trained using Adam optimizer for 60 epochs, with a batch size of 128 and an initial learning rate $3 \times {10^{ - 4}}$. 
The length of prompt tokens $C$ and the depth of prompt layers $J$ are set to 4 and 12, respectively.
We set the scale $s$ to 4 in the language branch and 0.1 in the vision branch. 
Rank-$k$ and mAP \cite{jiang2023cross} are utilized as the evaluation criteria. 


\subsection{Comparisons with State-of-the-art Methods}
\label{ssec: sota}

\noindent\textbf{Performance Comparisons.} In Table~\ref{table1} , CSKT outperforms other state-of-the-art methods on the largest dataset CUHK-PEDES, achieving 69.70\% Rank-1 and 62.4\% mAP.
The performance of CSKT achieving 58.90\% Rank-1 is superior to IRRA-CLIP,  while it lags behind CFine on ICFG-PEDES as Table~\ref{table2}. 
This observation suggests that CFine, as a more complex model having learnable 205M parameters, may have an advantage in fitting simpler datasets.
Experiments on the newly RSTPReid dataset in Table~\ref{table3} demonstrate that CSKT also surpasses SOTA  with the rise of +3.70\%, +3.03\% on Rank-1 and mAP, respectively. 

\textbf{Parameters Comparisons and Computation Efficiency.} It is noteworthy  that CSKT only trains 12M parameters as indicated in Table~\ref{table4}, which is merely 8.0\% learnable parameters compared to IRRA-CLIP (full-tuning CLIP) and 5.9\% compared with CFine.
Meanwhile, CSKT is computationally more efficient in training time/step : CSKT(0.107s) $<$ IRRA-CLIP(0.109s) $<$ CFine (0.418s).

\begin{figure}[tb]
\centerline{\includegraphics[scale=0.25]{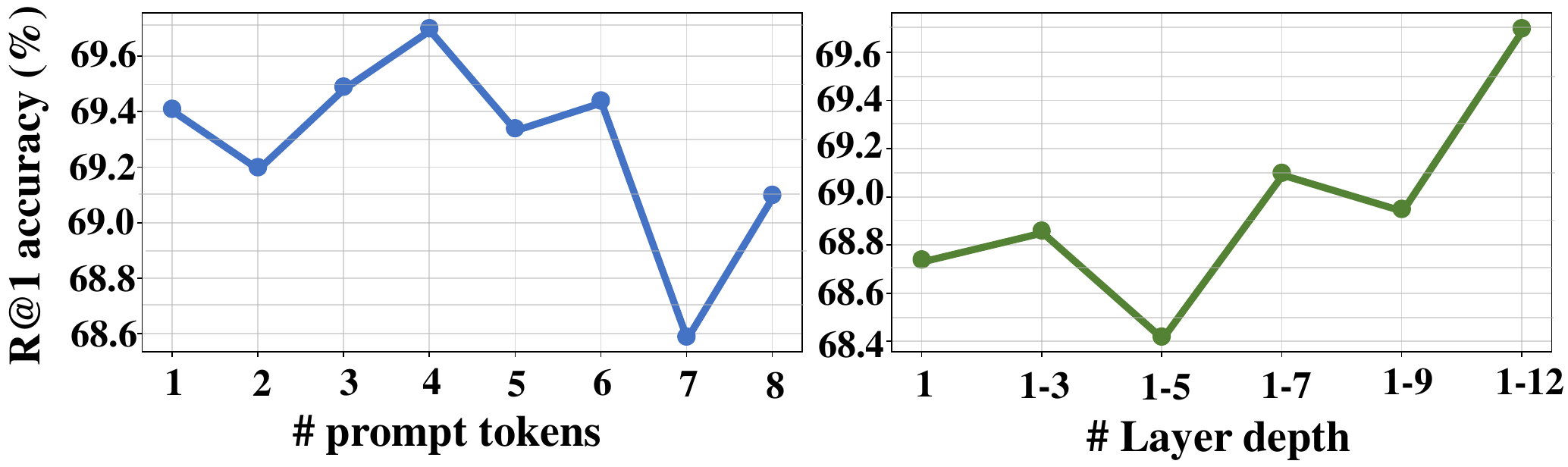}}
 \vspace{-3mm}
\caption{Ablation on prompt length and prompt depth.}
\label{fig:res}
 \vspace{-2mm}
\end{figure}
\begin{table}[tb]
\small
\centering
\tabcolsep=5pt
\renewcommand\arraystretch{1.1}
\caption{Comparison with other methods on CUHK-PEDES. The content in the left column denotes whether using  CLIP.}
\vspace{-2mm}
\resizebox{0.37\textwidth}{!}{%
\begin{tabular}{c|l|cccc}
\hline
                                               & Method    & R@1  & R@5  & R@10   & mAP  \\
\hline
\multirow{9}{*}{\rotatebox{90}{w/o CLIP}}      
                                            & CMPM/C \cite{zhang2018deep}      & 49.37     & 71.69      & 79.27      & -        \\
                                               & ViTAA \cite{wang2020vitaa}       & 55.97     & 75.84      & 83.52      & -      \\
                                               & DSSL \cite{zhu2021dssl}          & 59.98     & 80.41      & 87.56      & -      \\
                                               & SSAN \cite{ding2021semantically}         & 61.37     & 80.15      & 86.73      & -      \\   
                                               & SAF \cite{li2022learning}        & 64.13     & 82.62      & 88.40      & 58.61  \\
                                               & TIPCB \cite{chen2022tipcb}       & 64.26     & 83.19      & 89.10      & -      \\
                                               & AXM-Net \cite{farooq2022axm}     & 64.44     & 80.52      & 86.77      & 58.73  \\
                                               & LGUR \cite{shao2022learning}     & 65.25     & 83.12      & 89.00      & -  \\
                                               & IVT \cite{shu2022see}            & 65.59     & 83.11      & 89.21      & -  \\
\hline\hline
\multirow{4}{*}{\rotatebox{90}{w/ CLIP}}        & Han et al. \cite{han2021text}      & 64.08     & 81.73      & 88.19      & 60.08  \\

                                               & CFine \cite{yan2022clip}         & 69.57     & 85.93      & 91.15      & -  \\
                                               & IRRA-CLIP \cite{jiang2023cross} & 68.19     &86.47      & 91.47      & 61.12           \\
                                               & \textbf{CSKT (Ours)}     & \textbf{69.70}    & \textbf{86.92}    & \textbf{91.8}    & \textbf{62.74} \\
\hline
\end{tabular}}
\label{table1}
\vspace{-2mm}
\end{table}

\begin{table}[tb]
\small
\centering
\tabcolsep=5pt
\renewcommand\arraystretch{1.1}
\caption{Comparison on ICFG-PEDES.}
\vspace{-2mm}
\resizebox{0.38\textwidth}{!}{%
\begin{tabular}{c|l|cccc}
\hline
                                               & Method    & R@1  & R@5  & R@10   & mAP \\
\hline
\multirow{7}{*}{\rotatebox{90}{w/o CLIP}}                                                                                           & CMPM/C \cite{zhang2018deep}      & 
                                              43.51     & 65.44     & 74.26      & -  \\
                                               & ViTAA \cite{wang2020vitaa}       & 50.98     & 68.79     & 75.78      & -      \\
                                               & SSAN \cite{ding2021semantically} & 54.23     & 72.63     & 79.53      & -      \\
                                               & SAF \cite{li2022learning}        & 54.86     & 72.13     & 79.13      & 32.76    \\
                                               & TIPCB \cite{chen2022tipcb}       & 54.96     & 74.72     & 81.89      & -      \\
                                             & IVT \cite{shu2022see}            & 56.04     & 73.60     & 80.22      & -  \\
                                               & LGUR \cite{shao2022learning}     & 59.02     & 75.32     & 81.56      & -  \\
\hline\hline
\multirow{3}{*}{\rotatebox{90}{w/ CLIP}}
                                               & CFine \cite{yan2022clip}         & \textbf{60.83}    & 76.55      & 82.42      & -  \\
                                               & IRRA-CLIP \cite{jiang2023cross} & 56.74     & 75.72      & 82.26      & 31.84           \\
                                               & \textbf{CSKT (Ours)}     & 58.90    & \textbf{77.31}    & \textbf{83.56}    & \textbf{33.87} \\
\hline
\end{tabular}}
\label{table2}
\end{table}

\begin{table}[tb]
\small
\centering
\tabcolsep=5pt
\renewcommand\arraystretch{1.1}
\caption{Comparison on RSTPReid.}
\vspace{-2mm}
\resizebox{0.38\textwidth}{!}{%
\begin{tabular}{c|l|cccc}
\hline
                                               & Method    & R@1  & R@5  & R@10   & mAP \\
\hline
\multirow{4}{*}{\rotatebox{90}{w/o CLIP}}       & DSSL \cite{zhu2021dssl}          & 32.43                                                  & 55.08      & 63.19      & -      \\
                                               & SSAN \cite{ding2021semantically} & 43.50     & 67.80      & 77.15      & -      \\
                                               & SAF \cite{li2022learning}        & 44.05     & 67.30      & 76.25      & 36.81    \\
                                               & IVT \cite{shu2022see}            & 46.70     & 70.00     & 78.80      & -  \\

\hline\hline
\multirow{3}{*}{\rotatebox{90}{w/ CLIP}}       
                                               & CFine \cite{yan2022clip}         & 50.55     & 72.50     & 81.60      & -  \\
                                               & IRRA-CLIP \cite{jiang2023cross} & 54.05     & 80.70     & 88.00	& 43.41        \\
                                               & \textbf{CSKT (Ours)}     & \textbf{57.75}    & \textbf{81.30}    & \textbf{88.35}    & \textbf{46.43} \\
\hline
\end{tabular}}
\label{table3}
\vspace{-2mm}
\end{table}

\begin{table}[tb]
\small
\centering
\tabcolsep=3.5pt
\caption{Comparison about the size of fine-tuning parameters on training stage and R@1 on  CUHK-PEDES.}
\vspace{-2mm}
\renewcommand\arraystretch{1.1}
\resizebox{0.45\textwidth}{!}{%
\begin{tabular}{l|c|c|c|c}
\hline
                                                Method    & \#params  & \makecell[c]{Trainable\\Params (\%)}   & Traing Time/step (s)  & R@1  \\
\hline
     
                                                CFine \cite{yan2022clip}   & 205M  & 100 & 0.418 & 69.57\\
                                                IRRA-CLIP \cite{jiang2023cross} &  150M &100  & 0.109 &  68.19       \\
                                                \textbf{CSKT (Ours)}     & \textbf{12M}  & \textbf{7.4}  &   \textbf{0.107} & \textbf{69.70}  \\
\hline
\end{tabular}}

\label{table4}
\vspace{-2mm}
\end{table}

\subsection{Ablation Studies}
\label{ssec: ablate}

\begin{table}[!h]
\centering
\tabcolsep=3.5pt
\caption{Ablation study on each component of CSKT.}
\vspace{-2mm}
\renewcommand\arraystretch{1.1} 
  \resizebox{0.48\textwidth}{!}{%
  \begin{tabular}{c|l|ccc|ccc}
  \toprule
  \multirow{2}{*}{No.} &\multirow{2}{*}{Methods} &\multicolumn{3}{c|}{Components} &\multicolumn{3}{c}{CUHK-PEDES}  \\ 
  \cline{3-8}
       &                                         &UPT & BPT  &DAT      &R@1  &R@5  &R@10 \\ 
  \hline
  0    &Zero-shot CLIP                               &          &          &            &12.61	&27.08	&35.48    \\
  1    &+UPT &     \checkmark     &          &            &59.29   &79.65   &86.47     \\
  2    &+BPT                                     & &      \checkmark    &            &62.82	& 83.53 &	89.44	     \\    
  3    &+DAT                                     &          &          &\checkmark  &66.23   &85.51   &91.23   \\
  4    &+BPT+DAT (ours CSKT)                                   & &\checkmark&\checkmark  &\textbf{69.70}&\textbf{86.92}&\textbf{91.80}\\
  \bottomrule
  \end{tabular}%
  }
  \label{table5}
\vspace{-5mm}
  \end{table}

\textbf{Analysis on Components } We verify the effects of two main components: Bidirectional Prompts Transferring (BPT) and Dual Adapters Transferring (DAT) in Table~\ref{table5}. We adopt the Zero-shot CLIP-ViT-B/16 model as our baseline (\emph{No.0}). 
Additionally, we employ Unidirectional Prompts Transferring (UPT) (\emph{No.1}), similar to MaPLe, for  cross-modal prompting from language to vision branch. 
\emph{No.3} and \emph{No.4} illustrate the significant advantages of both BPT and DAT over the zero-shot CLIP baseline, with gains of 50.21\% and 48.34\% at Rank-1 respectively, indicating the superior effectiveness of PETL in TPR.
Compared to UPT, BPT improves the Rank-1/5/10 by 3.53\%, 3.88\% and 2.97\%, which infers that bidirectional \emph{dependent} strategy can promote tighter coupling between vision and language prompts to ensure mutual synergy and learn more dependent represatations that facilitate cross-modal feature alignment. 
When BPT and DAT are integrated in CSKT, Rank-1 improves by 6.88\% (\emph{No.2}) and 3.47\% (\emph{No.3}) compared to using them independently. 
This demonstrates that BPT in the input feature space of MHSA, and DAT in the output feature space of MHSA can work synergistically to carry out knowledge transfer in different positions of transformer.

\textbf{Analysis on Parameters.}
 Figure~\ref{fig:res}(\textbf{\emph{left}}) shows that when the length of prompt tokens increases to 4, CSKT achieves the best performance. Then, the accuracy decreases when exceeding this point, indicating over-fitting which inherently degrades the performance. 
Figure~\ref{fig:res}(\textbf{\emph{right}}) indicates that CSKT achieves maximum performance at the depth of 12.


\section{CONCLUSION}
\label{sec:typestyle}
In this paper, we introduce a simple and effective method CSKT for TPR, which is the first time that PETL
is successfully introduced in this task. BPT is designed to promote feature alignment and explore knowledge on feature input side. Furthermore, DAT is incorporated to transfer knowledge on output side of MHA within CLIP. The experimental results verify the superiority of CSKT when the trainable parameters only occupy 7.4\% of the entire model.

\section{ACKNOWLEDGMENT}
\label{sec:acknowledge}
This work was partly supported by the Special Foundations for the Development of Strategic Emerging Industries of Shenzhen(Nos.JSGG20211108092812020 \& JCYJ20200109-143035495).


\clearpage
\vfill\pagebreak
\bibliographystyle{IEEEbib}
\balance
\bibliography{strings,refs}

\end{document}